\documentclass{article}

  \usepackage[preprint,nonatbib]{neurips_2020}

\usepackage[utf8]{inputenc} 
\usepackage[T1]{fontenc}    
\usepackage{hyperref}       
\usepackage{url}            
\usepackage{booktabs}       
\usepackage{amsfonts}       
\usepackage{nicefrac}       
\usepackage{microtype}      
\usepackage{amsmath}
\usepackage{xcolor}
\usepackage{cancel}
\newcommand*{\var}{\text{Var} }
\newcommand*{\ex}{\mathbb{E} }
\usepackage{multirow}
\usepackage{graphicx}
\usepackage{etoc}

\title{Inner Ensemble Networks: Average Ensemble as an Effective Regularizer}

\author{%
  Abduallah Mohamed\textsuperscript{1}\And
  Muhammed Mohaimin Sadiq\textsuperscript{1}\And
  Ehab AlBadawy\textsuperscript{2}\And
   \AND
   Mohamed Elhoseiny\textsuperscript{3,*} \And
   Christian Claudel\textsuperscript{1,*} \AND
   \textsuperscript{1}UT Austin \And
   \textsuperscript{2}UAlbany \And
   \textsuperscript{3}KAUST, Stanford \And
  \textsuperscript{*} Equal Advising
}

\begin{document}
\etocdepthtag.toc{mtsection}
\etocsettagdepth{mtsection}{subsection}
\etocsettagdepth{mtappendix}{none}

\maketitle

\begin{abstract}
We introduce Inner Ensemble Networks (IENs) which reduce the variance within the neural network itself without an increase in the model complexity. IENs utilize ensemble parameters during the training phase to reduce the network variance. While in the testing phase, these parameters are removed without a change in the enhanced performance. IENs reduce the variance of an ordinary deep model by a factor of $1/m^{L-1}$, where $m$ is the number of inner ensembles and $L$ is the depth of the model. Also, we show empirically and theoretically that IENs lead to a greater variance reduction in comparison with other similar approaches such as dropout and maxout. Our results show a decrease of error rates between 1.7\% and 17.3\% in comparison with an ordinary deep model. We also show that IEN was preferred by Neural Architecture Search (NAS) methods over prior approaches. Code is available at \url{https://github.com/abduallahmohamed/inner_ensemble_nets}. 
\end{abstract}

\section{Introduction} 
\label{Introduction}

\begin{figure}[h]
\includegraphics[width=\linewidth]{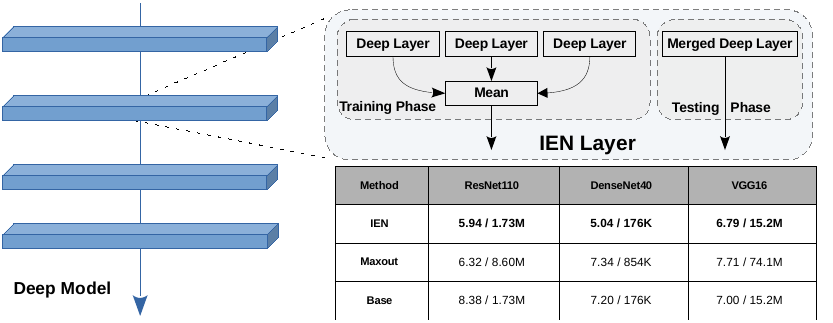}
\caption{During the training phase, inner ensemble weights are used to achieve better performance. In the testing phase, these extra weights are no longer necessary. The final model parameters count is the same as that of the original model without additional ensemble parameters. The table is accuracy/ parameters count on CIFAR-10 dataset.}
\label{fig:IENteaser}
\end{figure}

Ensemble learning~\cite{rokach2010ensemble,zhou2012ensemble} is based on a combination of predictions of multiple models trained over the same dataset or random subsets of the dataset to improve model performance. Ensemble methods have been widely used in deep learning~\cite{qiu2014ensemble,dietterich2000ensemble,drucker1994boosting,wasay2020mothernets} to improve overall model accuracy. Combining neural networks in ensembles is known to reduce the variance in the prediction. In other words, the ensemble of networks generalizes better to unseen data than a single network~\cite{krogh1995neural,geman1992neural,zhou2002ensembling}. However, there is a concern regarding the number of parameters of the final model, which involves the parameters of each constituent model. The large number of parameters in the final model decreases inference speed and increases model storage size, which can be problematic for some applications. In this paper, our objective is to find a solution that leads to an overall reduction in variance resulting from the ensembles, yet preserves the number of parameters of the underlying model.

We introduce Inner Ensemble Networks (IENs) to overcome the problem of the increased number of parameters in regular ensembles. IEN uses $m$ instances of the same layer and applies an average operation on the output of these layers as illustrated in Figure~\ref{fig:IENteaser}. IEN can be applied to each layer of a selected base model architecture. We hypothesize that the overall variance of the model can be decomposed into sub-variances within the model layers themselves. By using IENs, we reduce the sub-variances of each layer resulting in an overall variance reduction of the model. Therefore, we primarily explore the averaging method in our setting. Now we refer to regular ensembles as outer ensembles. The performance of IENs is between that of an ordinary deep model and an outer ensemble of multiple instances of the same model.

In this work, we discuss the theoretical aspects of variance reduction resulting from applying IENs. We show how to remove the excess weights used during the training and verify our approach empirically. We also contrast IEN variance reduction with closely related approaches such as dropout~\cite{JMLR:v15:srivastava14a} and maxout~\cite{goodfellow2013maxout} which gives a new perspective to these methods. When using IEN during the training phase, $m$ times the number of parameters in the base model are required to do an inner ensemble. This could be expensive if applied to Fully Connected (FC) layers. We show empirically that using IEN on Convolutional Neural Network (CNN) layers alone results in a performance that exceeds that of a base model. This makes the training cheaper because of the shared parameters property of CNNs. We also show that using IEN on both CNNs and Fully Connected (FC) layers similarly results in a performance which exceeds that of base models, but at the cost of increased training time. Lastly, we extend IEN as a Neural Architecture Search (NAS)~\cite{DBLP:journals/corr/ZophL16} cell. We applied NAS searching algorithms to discover new cells. The discovered cells show a preference for using IEN as a cell operator.
\section{Related work}
Ensemble methods are meta-algorithms that combine a set of independently trained networks into one predictive model in order to improve performance. Other methods include bagging and boosting which reduce predictive variance and bias~\cite{opitz1999popular}. Averaging the results of several models is one of the most common approaches and is extensively adopted in modern deep learning practice~\cite{russakovsky2015imagenet, rajpurkar2016squad}. One variation is to create ensembles of the same model at different training epochs~\cite{qiu2014ensemble}.~\cite{abbasian2013inner} proposed using inner ensembles to make learning decisions within the learning algorithms of classic machine learning techniques. Other approaches such as~\cite{ba2014deep} used a model compression technique, in which they trained a shallow network using an ensemble of networks. Yet, it has a high cost of first training the ensembles and then the actual shallow model unlike IENs which train a single model.~\cite{10.1007/978-3-319-54184-6_13} developed a method that averages the output layer of a model by encouraging diversity in subsets of the output layer neurons. A main concern is the complexity of the method which is not a simple plug-in method like IENs. In IENs you simply replace the CNN or FC layers with IEN layers without the need to change anything else. Another concern in~\cite{10.1007/978-3-319-54184-6_13} is that it forces a specific loss function unlike IENs which is agnostic from the choice of a loss function. Though inspired by ensemble methods that average independent networks, the proposed IEN structure is fundamentally different in that a) it only needs the extra ensemble weights during the training phase and b) it trains the ensembles jointly in a single model.

We could state that the idea of IEN existed implicitly in previous methods, starting from maxout~\cite{goodfellow2013maxout}. Maxout replicates deep layers and takes the maximum of the response coming from these replicas. Maxout can be considered as an ensemble that uses max instead of average. Yet, we show both theoretically and empirically that maxout is not a desirable option and IEN is a better alternative. Dropout~\cite{JMLR:v15:srivastava14a,NIPS2013_4878} can be seen as a geometric mean of several sub-networks within a deep model. We believe that IEN aligns with dropout as they are both only applied during the training phase. However, IEN can be applied on every layer of the deep model unlike dropout. Dropout was originally designed to be applied on FC layers. When it comes to CNNs, dropout is applied carefully~\cite{park2016analysis} and cannot be utilized to the fullest extent. We also show theoretically IEN variance reduction is greater than that of dropout when $m\geq 2$.

\section{Inner Ensemble Nets (IENs)}
Let us assume a deep model with $L$ layers. The $l$th layer has a response $\mathbf{y}_l$ from an input $\mathbf{x}_l$ multiplied by a learnable weight matrix $\mathbf{w}_l$, where $\mathbf{y}_l,\mathbf{x}_l \text{ and } \mathbf{w}_l$ are all tensors. An ordinary deep model layer can be formulated as: 
\begin{equation}\textbf{y}_l = \textbf{w}_l \textbf{x}_l\end{equation}
Thus, an IEN deep layer with $m$ inner ensembles can be defined as:
\begin{equation}\textbf{y}_l^{\text{IEN}}=\frac{1}{m} \sum_{i \in m} \textbf{w}_l^i \textbf{x}_l \end{equation}
Where  $\textbf{w}_l^i  =\{w^i_{kj}: k\in n, j \in d\} \in \mathbb{R}^{n\times d}$, $\textbf{x}_l =\{x_i: i \in d\} \in \mathbb{R}^{d\times1}$ and $\textbf{y}_l^{\text{IEN}} =\{y_i: i \in n\} \in \mathbb{R}^{n\times1}$. We notice that IEN trains a single model with multiple inner ensembles. Though IEN uses $m$ extra parameters during the training phase, the training time does not scale as a multiplier of $m$. This makes IEN an economical regularization method (Appendix~\ref{app:traintimemodelsize}).

\section{IEN theoretical analysis }
In this section, we start with analyzing IEN variance response. We discuss the method of removing the ensemble weights. Then we contrast maxout and dropout variance response with that of IEN.
\subsection{IEN variance response}
In this section we show that IEN ensembles of size $m \in \mathbb{Z}^{>1}$ decrease the overall variance of a deep net by a factor of:
\begin{equation}
    \frac{1}{m^{L-1}} \text{, where $L$ is the number of layers in the network.}
\label{eq:IEN_fact}
\end{equation}
We end up with a new initialization that prevents the explosion or degradation of the gradients. To analyze a deep model variance that uses IEN we use the assumptions of~\cite{10.1109/ICCV.2015.123} and~\cite{pmlr-v9-glorot10a}: 
\begin{itemize}
    \item $\textbf{w}_l^i$ elements are initialized to be mutually independent with zero mean and a symmetric distribution sharing the same distribution variance.
    \item $\textbf{x}_l$ elements are initialized to be mutually independent sharing the same distribution with zero mean.
    \item $\textbf{w}_l$ and $\textbf{x}_l$ are independent from each other.
\end{itemize}
The variance of each $y_l^{\text{IEN}} \in \textbf{y}_l^{\text{IEN}} $is:
\begin{equation}\var[y_l^{\text{IEN}}]= n_l \var [\frac{1}{m}\sum_{i \in m} w_l^i x_l] = n_l \frac{1}{m^2} \var [\sum_{i \in m} w_l^i x_l] \end{equation}
Because $w_l^i$ and $x_l$ are independent, which implies they are uncorrelated, we can state: 
\begin{equation}\var[y_l^{\text{IEN}}]= n_l \frac{1}{m^2}\sum_{i \in m} \var [w_l^i x_l]= n_l \frac{1}{m^2} \var[x_l] \sum_{i \in m} \var[w^i_l ]\end{equation}
%
%
%
Because all elements of $w^i_l$ share the same distribution parameters, their variances are equal. We can define $\var[w^i_l] = \var[w_l]$ and by their independence: %
\begin{equation}\var[y_l^{\text{IEN}}] =  n_l \frac{1}{m} \var[x_l]  \var[w_l]\end{equation}

By using the the results and assumptions made at Appendix~\ref{app:vofnt} given that $x_l=f(y_{l-1})$, where $f$ is an activation function and $\beta^2$ is a the gain of $f$, we have: 
\begin{equation}\var[y_l^{\text{IEN}}] =  n_l \beta^2 \frac{1}{m} \var[w_l]  \var[y_{l-1}]\end{equation}

For cascaded $l$ layers and with the fact that $\var[y_1]$ represents the variance of the input layer, we arrive at: 
\begin{equation}
\label{eq:IEN}
\var[y_l^{\text{IEN}}] =   \var[y_1] (\prod_{l=2}^L \beta^2 n_l \frac{1}{m}  \var[w_l]) \end{equation}

An ordinary deep model variance response~\cite{10.1109/ICCV.2015.123} is: 
\begin{equation}
\var[y_L] = \var[y_1] (\prod_{l=2}^L \beta^2 n_l  \var[w_l]) 
\label{eq:normal}
\end{equation}
From equation \ref{eq:IEN} and \ref{eq:normal}, the variance of response using IEN is $
\frac{1}{m^{L-1}}$ times less than using an ordinary deep model. This states that going deeper using IEN layers or wider using $m$ will lead to better generalization of the model. We verified this empirically in the experiments section. We also want the multiplication in \ref{eq:IEN} to have a proper scalar in order to avoid reduction or magnification of the input signal, we want: 
\begin{equation} \frac{\beta^2 n_l}{m}  \var[w_l]  = 1, \forall l\end{equation}
So we need to initialize our weights to be: 
\begin{equation}
    w_l^i \sim \mathcal{N}(0,\frac{ m}{\beta^2 n})
\end{equation}
For simplicity, we use the same initialization for the first layer. 

\subsection{Reverting back to the original network size}
\label{sec:downsize}
Though the IEN reduces the variance of the network, we still have the burden of the extra parameters. Because of the training mechanism of IEN, all $m$ ensembles receive the same error signal. This results in weights that are close to each other in value, but not exactly the same because of the initialization. The most straight forward approach is to average these weights. The final weight $\widetilde{\textbf{w}}_l$ of IEN becomes:
\begin{equation}
\widetilde{\textbf{w}}_l = \frac{1}{m} \sum_{i\in m}  \textbf{w}^i_l
\label{eq:inv_var}
\end{equation}
Thus, our IEN layer becomes: $
\textbf{y}_l^{\text{IEN}}=\widetilde{\textbf{w}}_l \textbf{x}_l$ which has the exact same number of parameters as before using the IEN, but with the added benefit of variance reduction. Hence, we were able to revert back to the original model size for the inference stage, yet we utilized the extra parameters of IEN during the training phase. Natural questions arise regarding whether the method of averaging the weights is applicable to maxout and whether it is applicable to $m$ separately trained models. The answer to both is no. We verified these findings empirically in the experiments section.

\subsection{Connection with dropout}
Dropout~\cite{JMLR:v15:srivastava14a} is considered as a geometric mean of several small networks~\cite{NIPS2013_4878}. Here we try to connect dropout with IEN and contrast the variance performance. We want to note that we do not introduce IEN as an alternative for dropout. It is simply a tool to be used alongside other deep learning regularization tools. Dropout can be defined as:
\begin{equation}
y_l^{\text{dropout}}= \delta_l w_l x_l
\end{equation}
Where $\delta_l \sim \text{Bernoulli}(p)$ of probability $p$. From  B, the variance of a dropout layers is upper bounded as follows:
\begin{equation}
\var[y_L^{\text{dropout}}] \leq \var[y_1] (\prod_{l=2}^L \beta^2 n_l  \frac{1}{2}  \var[w_l]) 
\end{equation}
Thus, dropout decreases the variance response at most: 
\begin{equation}
\leq\frac{1}{2^{L-1}} \text{ times less than using an ordinary deep model.}
\label{eq:dropout_fact}    
\end{equation}
Comparing equations~\ref{eq:dropout_fact} and~\ref{eq:IEN_fact}, we conclude that IEN outperforms dropout if used alone when $m > 2$.
\subsection{Connection with maxout}
Maxout~\cite{goodfellow2013maxout} is a universal approximator. It shares a common setting with IEN in terms of having multiple replicas of the weights and it selects the max response out of these replicas. One disadvantage of maxout is that the required number of parameters in both training and inference stays the same, unlike IEN with a finale averaged weight that solves the problem for inference as stated in section~\ref{sec:downsize}. A maxout layer is defined as:
\begin{equation}
\textbf{y}_l^{\text{maxout}}= \max\{\textbf{w}_l^i \textbf{x}_l\}_i^m
\end{equation}
Because the usage of $\max$ makes the analysis of variance much harder, we provide three views for the variance of maxout. The first one is a general upper bound (Appendix~\ref{app:maxupper}) on the variance of maxout, which is not tight, but is helpful in our analysis. The bound is:
\begin{equation}
\var[y_L^{\text{maxout}}] \leq \var[y_1] (\prod_{l=2}^L n_l m \beta^2   \var[w_l]) 
\end{equation} 
which suggests that maxout might result in an explosion in the variance by a factor of
\begin{equation}
m^{L-1} \text{ more than using an ordinary deep model.}
 \label{eq:maxoutupper}
\end{equation} 
in comparison with both IEN and even an ordinary deep model. We observe this behavior in Table~\ref{tb:mainresults}. 

The second bound of maxout states the lowest possible variance~\cite{ding2015multiple} (Appendix~\ref{app:maxlower}) it can achieve under an assumption that $(w_l^i x_l) \sim \mathcal{N}(0,1)$ and a linear activation function is: 
\begin{equation}
\var[y_l^{\text{maxout}}] \geq n_l \frac{c}{\log{m}}, c>0
\label{eq:maxoutlower}
\end{equation}

The IEN variance reduction gain under the later assumptions and by using equation~\ref{eq:IEN_fact} is: 
\begin{equation}
\var[y_l^{\text{IEN}}] = n_l \frac{1}{m}
\label{eq:IEN_fact_single}
\end{equation}

This maxout bound shows that as the number of ensembles $m$ increases, the variance decreases, controlled by the value of $c$. The issue is that this bound is only for a single-layer model and it is difficult to extend it to multi-layer models. Yet it gives some insights. Depending on the value of $c$ in equation~\ref{eq:maxoutlower}, the IEN could perform better or worse. We highlight that these findings are limited and only valid for the aforementioned assumptions.

The last bound is the asymptotic bound by using extreme value theorem~\cite{leadbetter1988extremal} (Appendix\ref{app:maxasymp}). When $m$ is sufficiently large under the assumptions that $(w^i_l x_l)\sim \mathcal{N}(0,\sigma^2)$ and $f$ is a linear activation function, the bound is: 
\begin{equation}
\var[y_l^{\text{maxout}}] \approx n_l \frac{\pi^2}{6} \frac{\sigma^2}{2 \ln m}
\label{eq:maxoutasymp}
\end{equation}
Suggesting that the asymptotic maxout bound is:
\begin{equation}
 \approx  \frac{\pi^2}{12 \ln (m)} \text{ times less than ordinary deep model}
\end{equation}

\subsection{Summary of the variance gains}
\begin{figure}[h!]
\includegraphics[width=\linewidth]{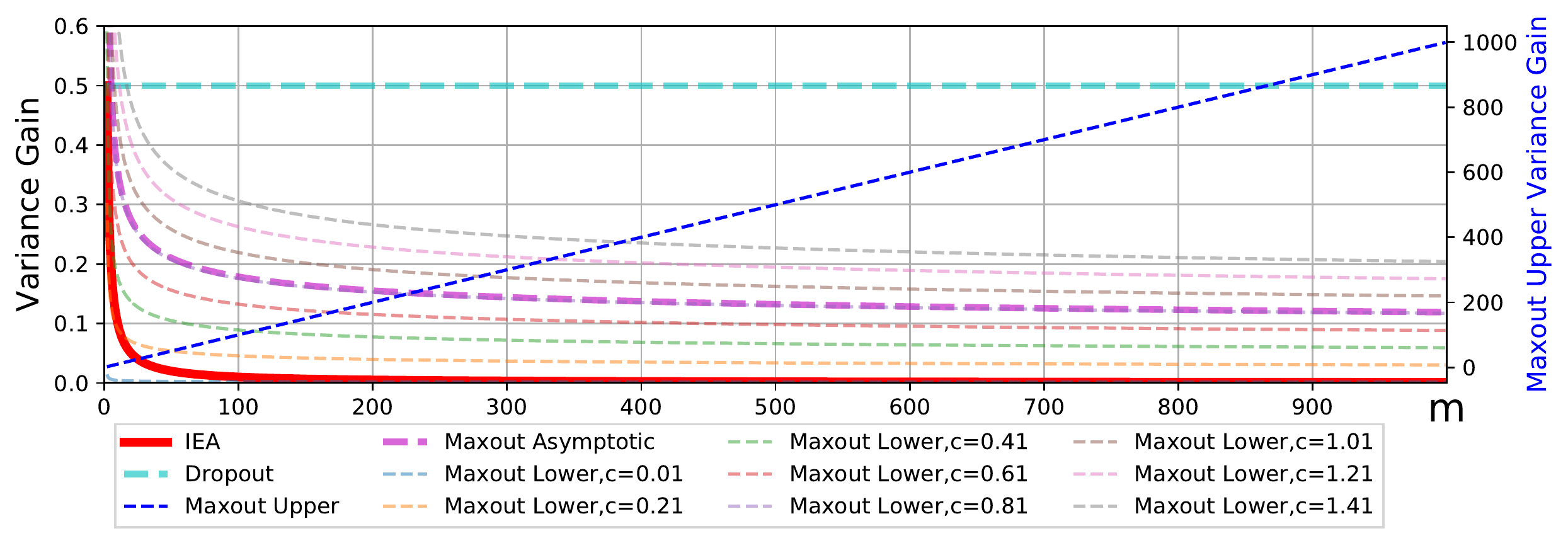}
\caption{The variance gain of IEN, maxout and dropout versus the number of ensembles in a single linear layer model. Please note that maxout upper variance gain has a different scale. }
\label{fig:var}
\end{figure}

Figure~\ref{fig:var} summarizes the variance behaviour in IEN, dropout and maxout. We assume a model with one layer to have common settings that aligns with all derived bounds and $f$ is a linear activation function. We used equation~\ref{eq:IEN_fact_single} for IEN, and equation~\ref{eq:dropout_fact} for dropout and equations~\ref{eq:maxoutlower},~\ref{eq:maxoutasymp} and ~\ref{eq:maxoutupper} for maxout. We observe that IEN variance reduction surpasses all previous methods and the different maxout variance bounds except when $c$ is extremely small in maxout lower variance bound. We also observe that the maxout upper bound explodes compared to IEN and dropout. Dropout variance reduction remains the same.

\section{Experiments \& discussions}
We wanted to verify that IEN generalized the deep model better than other approaches. For this we chose three well-known deep models based on the extent of their residual connections, starting from VGG~\cite{simonyan2014very} which has no residual connections, then ResNet~\cite{DBLP:journals/corr/HeZRS15} which has one residual connection, and finally DenseNet~\cite{DBLP:journals/corr/HuangLW16a} which is a fully residual connected net. We believe that benchmarking against these models will give an overview that be can generalized to other architectures. We use both CIFAR-10 and CIFAR-100~\cite{krizhevsky2009learning} for all of our experiments as the former shows the performance on a small classification task and the latter on a larger classification task. We also choose to experiment on image classification models as they use both CNN and FC layers. We used the \textit{original hyper-parameters} of the models in the experiments\textit{ without any tuning in favor of our method}.

\begin{table}[ht]
\caption{Error rate mean and std of IEN, maxout and the original model design on different deep model architectures. The lower, the better. The subscript $\widetilde{\textbf{w}}$ indicates results using the weight downsizing method from section~\ref{sec:downsize}. \textit{+FC} stands for IEN applied on both FC and CNN layers. Maxout results are only on CNN layers. The \textcolor{blue}{blue} colored models have the exact same number of parameters as the corresponding base models. The rest have the same number of parameters that scales with $m$. IEN and maxout have $m=4$.}
\centering
\scriptsize
\begin{tabular}{llllllll}
        Dataset                   &        & ResNet56 & ResNet110 & DenseNet40-12 & DenseNet100-12 & VGG16 & VGG19  \\ 
\hline
\multirow{7}{*}{CIFAR-10}  &\textcolor{blue}{$\text{IEN}_{\widetilde{\textbf{w}}}$}    & \textbf{6.93$\pm$0.26}    & \textbf{6.27$\pm$0.32}     & 7.50 $\pm$ 0.18        & \textbf{5.22$\pm$0.19}          & \textbf{6.80$\pm$0.18 }& \textbf{6.79$\pm$0.30 } \\
                           & IEN  & \textbf{6.93$\pm$0.26}    & \textbf{6.28$\pm$0.32}    & 7.5 $\pm$ 0.18         & \textbf{5.22$\pm$0.19 }          & \textbf{6.79$\pm$0.19} & \textbf{6.79$\pm$0.31 } \\
                           &\textcolor{blue}{ $\text{IEN+FC}_{\widetilde{\textbf{w}}}$}    & \textbf{6.49$\pm$0.04}    &\textbf{5.94$\pm$0.26}     &\textbf{6.89$\pm$0.20 }        &  \textbf{5.04$\pm$0.07 }       & 6.85$\pm$0.18 & 7.15$\pm$0.27  \\ 
\cline{2-2}
                           & \textcolor{blue}{$\text{Maxout}_{\widetilde{\textbf{w}}}$}    & 89.99$\pm$0    & 90$\pm$0     & 89.70$\pm$0.51         &  90$\pm$0          & 90$\pm$0 & 90$\pm$0  \\
                           & Maxout  & 7.98$\pm$0.32    & 6.32$\pm$0.12     & 7.34$\pm$0.155         & $6.16\pm$0.54          & 7.71$\pm$0.15 & 8.16$\pm$0.18  \\ 
\cline{2-2}
                           &\textcolor{blue}{ $\text{Base}_{\widetilde{\textbf{w}}}$}   & 89.99    & 90     & 90         & 90          & 90 & 90  \\
                           & \textcolor{blue}{Base} & 8.38$\pm$1.2    & 6.38$\pm$0.48     & 7.20$\pm$0.15         & 5.6$\pm$0.12          & 7.00$\pm$0.08 & 7.02$\pm$0.08  \\ 
\hline
\multirow{7}{*}{CIFAR-100} & \textcolor{blue}{$\text{IEN}_{\widetilde{\textbf{w}}}$}    &\textbf{29.34$\pm$0.49 }   & 28.17$\pm$0.11     & \textbf{29.37$\pm$0.30}         & \textbf{23.76$\pm$0.45}          & \textbf{28.49$\pm$0.39 }& \textbf{29.40$\pm$0.22 } \\
                           & IEN  & \textbf{29.34$\pm$0.49}    & 28.16$\pm$0.10     & \textbf{29.37$\pm$0.30 }        & \textbf{23.76$\pm$0.45}          & \textbf{28.50$\pm$0.39} &\textbf{29.41$\pm$0.23 } \\
                           & \textcolor{blue}{$\text{IEN+FC}_{\widetilde{\textbf{w}}}$}   &\textbf{28.20$\pm$0.19}    & \textbf{27.34$\pm$0.28 }    & \textbf{29.76$\pm$0.19}         & \textbf{23.67$\pm$0.37}          & 29.33$\pm$0.10 & 31.81$\pm$0.19  \\ 
\cline{2-2}
                           & \textcolor{blue}{$\text{Maxout}_{\widetilde{\textbf{w}}}$}    & 99.01$\pm$0.01    & 99$\pm$0     & 99$\pm$  0       & 99 $\pm$  0          & 99$\pm$0 & 99$\pm$0  \\
                           & Maxout    & 31.8$\pm$1.69    & 29.47$\pm$0.84     & 30.49$\pm$0.75         & 28.88$\pm$5.72          & 31.68$\pm$0.65 & 34.32$\pm$0.30  \\ 
\cline{2-2}
                           & \textcolor{blue}{$\text{Base}_{\widetilde{\textbf{w}}}$}   & 98.96    & 99.14     & 99         & 90          & 99 & 99  \\
                           & \textcolor{blue}{Base} & 29.97$\pm$0.71    & 27.83$\pm$0.64     & 29.85$\pm$0.39         & 24.01$\pm$0.14          & 29.27$\pm$0.27 & 30.92$\pm$0.49  \\ 

\bottomrule
\end{tabular}
\label{tb:mainresults}
\end{table}

\subsection{IEN behavior analysis}
\textbf{Where to apply IEN:} IEN can be applied to both CNN and FC layers. The question that arises concerns the cost of training time which is correlated with the number of parameters. Another question is whether it is better to apply IEN to just CNN layers or both CNN and FC layers. From Table~\ref{tb:mainresults}, applying IEN to CNN layers only, \textcolor{blue}{$\text{IEN}_{\widetilde{\textbf{w}}}$}, yields better results than the base model as observed for ResNet, VGG an DenseNet. Also, it might not be enough to apply it only to CNN layers, as observed in DenseNet40-12 results. Yet, when it is applied to FC layers, \textcolor{blue}{ $\text{IEN+FC}_{\widetilde{\textbf{w}}}$}, too it will yield better results than the base model. So, the model designer has a choice. If IEN is applied to CNN layers only, it will save training time and enhance the performance. If applied to both CNN and FC layers it will increase the training time, but will result in a better performance than the former.

\textbf{Weight downsizing property:} We denote the downsized model using the method from section~\ref{sec:downsize} with subscript $\widetilde{\textbf{w}}$. We observe from Table~\ref{tb:mainresults} that downsized IEN models which have the same number of parameters as the base model do have the same performance as the one with the full ensemble parameters. This property was verified across different architectures as shown in the same table. Also, it holds for applying IEN on CNN or FC layers. We also wanted to verify if this property can be extended to the base model or the maxout model. We applied $\widetilde{\textbf{w}}$ on both of them and as observed from Table~\ref{tb:mainresults} their performance was extremely poor and close to random guessing. We observe that the weight downsizing property was only successful in IEN for two reasons. Firstly, the mean is a linear operator allowing the inner ensembles weights to receive the same error signal during the training phase. Secondly, the weights were initialized from the same distribution. This leads the weights to behave very similarly, resulting in closely related weights.

\textbf{Connection with the variance analysis:} From Table~\ref{tb:mainresults}, we observe that IEN outperforms previous methods such as maxout and the base model. We now recall from the variance reduction gain of IEN equation~\ref{eq:IEN_fact} that going deeper leads to better results. We notice this happens in both ResNet and DenseNet. Yet, it does not hold for VGG-16 and VGG-19 which is the same case for maxout and the base model. This suggests the necessity of residual connections as in~\cite{veit2016residual}. Aside from VGG results, our variance analysis of IEN aligns with the empirical findings. For maxout, we notice in some cases such as ResNet56 results for CIFAR-10, that it performs better than the base model. This is related to the maxout lower bound that was discussed earlier. However, in some cases such as ResNet56 for CIFAR-100, maxout performs far worse than the base model. This aligns too with the upper and asymptotic variance bounds of maxout. The dropout results are at Appendix~\ref{app:emprdropout}.

\begin{figure}[h]
\includegraphics[width=\linewidth]{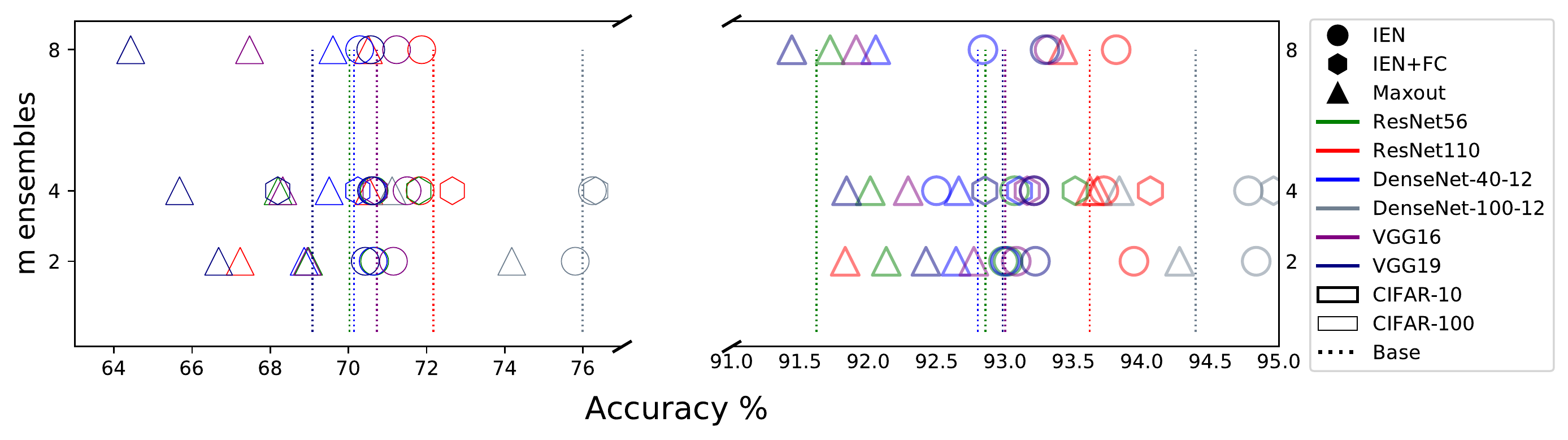}
\caption{The effect of changing $m$ versus model mean accuracy. IEN, maxout and base model accuracies are shown. CIFAR-10 results are the thin line shapes, while CIFAR-100 results are the thick line shapes. Colors represents different architectures. Dashed line is the base model mean accuracy.}
\label{fig:numberofm}
\end{figure}

\textbf{Effect of number of ensembles $m$: } 
We also wanted to study the effect of number of ensembles $m$ on accuracy of the models. Figure~\ref{fig:numberofm} illustrates this affect. We notice that IEN with $m=[2,4,8]$ outperforms maxout with the same $m$. Most of the time maxout performs worse than the base model. In a few cases such as CIFAR-10 ResNet56, maxout might perform better than the base model but not better than IEN. Most of the time IEN and IEN+FC outperform the base model. Lastly, using $m=[2,4]$ is much better than using $m=8$. This suggests that going higher with the number of inner ensembles is not desirable. This behavior aligns with the typical behavior of an outer ensemble~\cite{rosen1996ensemble}.
\subsection{Comparison with outer-ensemble}

We denote regular ensembles as outer ensembles to distinguish them from inner ensembles. From Table~\ref{tb:outerensemble}, an outer ensemble of IENs exceeds in performance an outer ensemble of base models and maxout models using the same number of parameters except in a few cases. From Table~\ref{tb:outerensemble} and Table~\ref{tb:mainresults} we notice that an outer ensemble of the base model outperforms single model IEN, but an IEN model is better than a single base model. This indicates that IEN performance stands in between a single regular model and an ensemble of the same model.
\begin{table}[h]
\caption{Results of outer-ensemble of the models. Each model was trained 3 times and the predictions were averaged. The metric reported is the mean of the error rate. The lower the better. The subscript $\widetilde{\textbf{w}}$ indicates results with the weight downsizing method from section~\ref{sec:downsize}. \textit{+FC} stands for IEN applied on Fully Connected layer. All IENs and Maxouts have $m=4$. The base model ensemble results are from 4 trained models.}
\centering
\scriptsize       
\begin{tabular}{llllllll}
         Dataset                  &      & ResNet56 & ResNet101 & DenseNet40-12 & DenseNet100-12 & VGG16 & VGG19  \\ 
\hline
\multirow{4}{*}{CIFAR-10}  & $\text{IEN}_{\widetilde{\textbf{w}}}$    & \textbf{5.49}    & 5.00     & 5.91         & \textbf{4.02}          & 5.68 & 5.71  \\
                           & $\text{IEN+FC}_{\widetilde{\textbf{w}}}$  & \textbf{5.05}    &\textbf{4.95}     & \textbf{5.21}         & \textbf{4.04}          & 5.70 & 6.02  \\
                           & Maxout  & 6.43    & 5.20     & 5.66        &   4.87        & 6.24 & 6.66  \\
                           & Base & 5.68    & 4.97      & 5.46         & 4.59         & \textbf{5.61} & \textbf{5.67}  \\ 
\hline
\multirow{4}{*}{CIFAR-100}  & $\text{IEN}_{\widetilde{\textbf{w}}}$    & 24.38    & 23.32     & 24.85         & \textbf{19.65}          & \textbf{24.60} & \textbf{25.33 } \\
                           & $\text{IEN+FC}_{\widetilde{\textbf{w}}}$  & \textbf{23.62}    & 23.03     & \textbf{24.44}         & \textbf{19.11}        & 25.44 & 27.44  \\
                           & Maxout  & 26.34    & 24.12     & 24.72         & 24.65          & 26.89 & 29.29  \\
                           & Base & 24.13    &\textbf{22.26}     & 24.85         & 19.86         & 24.76 & 26.02  \\ 
\bottomrule
\end{tabular}

\label{tb:outerensemble}
\end{table}

\subsection{Extension to NAS}
\begin{figure}[h]
\includegraphics[width=\linewidth]{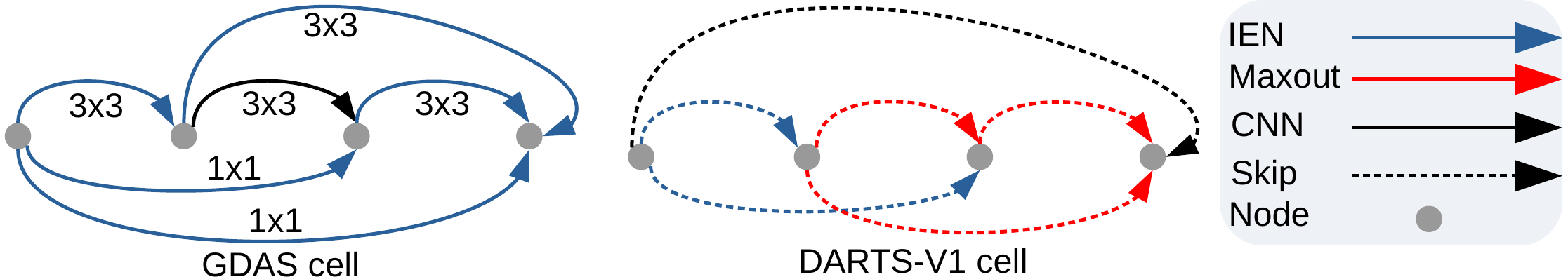}
\caption{Cells discovered using NAS DARTS-V1 and GDAS search algorithms. $1\times 1$ and $3\times 3$ are the CNN kernel sizes. IEN and maxout have $m=4$. The cells are acyclic graphs. The dashed arrows are skip connections. Colors represent different approaches.  }
\label{fig:nas}
\end{figure}
In order to make sure that IEN is preferred in deep model design, we tested it using NAS search algorithms. We introduced IEN and maxout as operators in both DARTS-V1~\cite{DBLP:journals/corr/abs-1806-09055} and GDAS~\cite{8953848} search algorithms besides regular CNNs. We ran NAS-Bench-201~\cite{dong2020nasbench201} on both DARTS and GDAS to discover the cells. We report results on CIFAR-10, CIFAR-100, and ImageNet-16-120. ImageNet-16-120 is a subset of ImageNet~\cite{ILSVRC15} that is used by NAS-Bench-201. From Figure~\ref{fig:nas} GDAS shows a high preference towards using IENs. DARTS-V1 resulted in a mix between maxout and IEN and used IEN in the early stage of the cell. This early usage of IEN might correlate with the importance of reducing the variance in the early layers of a deep model. Also, DARTS resulted in a skip-connect cell which is a known bias in DARTS algorithm~\cite{dong2020nasbench201}. Table~\ref{tb:NASresults} shows that our discovered cell which depends on IEN resulted in a better performance than cells that use regular CNNs.
\begin{table}[h]
\caption{Accuracies of four cell models based on NAS-Bench-201 configuration, using our discovered cells and the original NAS-Bench-201 cells.}
\small
\centering
\begin{tabular}{lllll}
NAS Search Method          &     Cell          & CIFAR-10 & CIFAR-100 & ImageNet-16-120  \\ 
\hline
\multirow{2}{*}{DARTS-V1}  & \textbf{Ours}          & \textbf{41.28}    & \textbf{17.36}     & \textbf{16.77}            \\
                           & NAS-Bench-201 Cell & 39.77    & 15.03     & 16.43            \\ 
\hline
\multirow{2}{*}{GDAS} & \textbf{Ours}           & \textbf{91.26$\pm$0.11}    & \textbf{72.68$\pm$0.12}      & \textbf{47.6$\pm$0.33}            \\
                           & NAS-Bench-201 Cell           & 89.89$\pm$0.08    & 71.34$\pm$0.04     & 41.59$\pm$ 1.33   \\
\bottomrule
\end{tabular}

\label{tb:NASresults}
\end{table}
\section{Conclusion}
We presented IENs which enhance the performance of a deep model by utilizing the concept of ensembles while keeping the model parameters free from the excess ensemble weights. We theoretically analyzed the variance reduction gain of IEN and other approaches with respect to ordinary deep model variance. We empirically showed that IENs outperformed other methods when applied on known deep model architectures. We analyzed IEN behavior versus that of outer ensembles and extended it to NAS with the discovery of new cells.

\clearpage



\bibliographystyle{abbrv}
\bibliography{ienbib}

\clearpage
\appendix
\section*{Appendix}

\etocdepthtag.toc{mtappendix}
\etocsettagdepth{mtsection}{none}
\etocsettagdepth{mtappendix}{subsection}
\tableofcontents
\clearpage
\section{Lemmas}
\label{app:Lemmas}
\subsection{Variance of the product of independent random variables}
\label{app:poirv}
if $y_1,y_2,...,y_n$ are independent random variables, then:
\begin{equation}
\var[y_1*y_2...y_n]= \prod_{i=1}^n (\var[y_i]+\ex[y_i]^2) - \prod_{i=1}^n \ex[y_i]^2 
\end{equation}

\subsection{Variance gain of a non-linear activation function}
\label{app:vofnt}
Let $y$ be a random variable, $x = f(y)$, where $f$ is a nonlinear activation function, one can state that:
\begin{equation}
\var[x] = \int_{-\infty}^{\infty} f(y)^2 \mathbb{P}(y) dy
\end{equation}
By using first order Taylor series expansion:
\begin{equation}
\var[x] \approx \int_{-\infty}^{\infty} (f(y_0)+f'(y_0)(y-y_0))^2 \mathbb{P}(y) dy
\end{equation}
Where $y_0 = 0$, then: 
\begin{equation}
\label{eq:gain}
\var[x] \approx \int_{-\infty}^{\infty} f(0)^2 \mathbb{P}(y) dy+  \int_{-\infty}^{\infty} f'(0)^2y^2 \mathbb{P}(y) dy +2 \int_{-\infty}^{\infty} f(0)f'(0)y \mathbb{P}(y) dy
\end{equation}
While Equation~\ref{eq:gain} is valid only for smoothness C1 functions it also works reasonably on non C1 functions like ReLU= $\max(0,y)$.
If we choose $x=\max(0,y)$ aka a ReLU function, we have: 
\begin{equation}
     x'(y)=\begin{cases}
       1&\quad\text{if } y>=0\\
       0&\quad\text{if } y<0\\
     \end{cases}
\end{equation}

By using Equation~\ref{eq:gain} and by assuming $x$ has zero mean with symmetric distribution:
\begin{equation}
\var[x] \approx  \int_{-\infty}^{\infty} f'(0)^2y^2 \mathbb{P}(y) dy  =  \int_{0}^{\infty}y^2 \mathbb{P}(y) dy 
\end{equation}
Thus, we can state: 
\begin{equation}
\var[x] \approx  \frac12  \int_{-\infty}^{\infty}y^2 \mathbb{P}(y) dy =  \beta^2 \var[y]
\end{equation}
We denote the gain $\frac12 $as $\beta^2$. Same procedure can be applied to different activation functions. A full table of different gains for commonly used deep model activation function can be found at~\cite{pytorchKHInit}.

Another approach for ReLU gain is~\cite{sgKHI}: 
$$
\var[x] = \int_{-\infty}^{\infty} x^2 \mathbb{P}(x) dx
$$
$$
\var[x] = \int_{-\infty}^{\infty} \max(0,y)^2 \mathbb{P}(y) dy
$$
$$
\var[x] = \int_{0}^{\infty} y^2 \mathbb{P}(y) dy
$$
$$
\var[x] =\frac{1}{2}* \int_{-\infty}^{\infty} y^2 \mathbb{P}(y) dy
$$
$$
\var[x] =\frac{1}{2}*\var[y] 
$$
$$
\var[x] =\beta^2*\var[y] 
$$

\section{Dropout variance response}
\label{app:dropoutderv}
$$\textbf{y}_l^{\text{dropout}}= \Delta_l \circ \textbf{w}_l \textbf{x}_l $$
Assumptions: 
\begin{itemize}
    \item $\textbf{w}_l  =\{w_{ji}: j\in n, i \in d\} \in \mathbb{R}^{n\times d}$ elements are initialized to be mutually independent with zero mean and a symmetric distribution
    \item $\textbf{x}_l =\{x_i: i \in d\} \in \mathbb{R}^{d\times1}$ elements are initialized to be mutually independent sharing the same distribution with zero mean
    \item $\Delta_l  =\{\delta_i: i \in n\} \in \mathbb{R}^{n\times1}$ elements are Bernoulli $\delta_l$ mutually independent random variables which are multiples element wise by $\textbf{w}_l $
    \item $\textbf{w}_l$, $\Delta_l$ and $\textbf{x}_l$ are independent from each other
    \item $\textbf{y}_l^{\text{dropout}} =\{y_i: i \in n\} \in \mathbb{R}^{n\times1}$ 
\end{itemize}
The variance of each $y_l^{\text{dropout}} \in \textbf{y}_l^{\text{dropout}} $, where $l$ is the layer number, is: 
\begin{equation}\var[y_l^{\text{dropout}}]= n_l \var[\delta_l w_l x_l] \end{equation}

By using the lemma in~\ref{app:poirv}:

\begin{equation}\var[y_l^{\text{dropout}}]=  n_l[ (\var[\delta_l]+\ex[\delta_l]^2) (\var[w_l]+\ex[w_l]^2) (\var[x_l]+\ex[x_l]^2) - (\ex[\delta_l]^2) (\ex[w_l]^2) (\ex[x_l]^2)]
 \end{equation}
\begin{equation}\quad\quad\quad\quad\quad=  n_l[ (\var[\delta_l]+\ex[\delta_l]^2) (\var[w_l]+\cancelto{0}{\ex[w_l]^2}) (\var[x_l]+ \cancelto{0}{\ex[x_l]^2}) - (\ex[\delta_l]^2) (\cancelto{0}{\ex[w_l]^2}) (\cancelto{0}{\ex[x_l]^2})]
 \end{equation}
 Thus,
\begin{equation}\var[y_l^{\text{dropout}}]= n_l (\var[\delta_l]+\ex[\delta_l]^2) \var[w_l] \var[x_l] \end{equation}

By using the the results and assumption made at ~\ref{app:vofnt} given that $x_l=f(y_{l-1})$, we have: 
\begin{equation}\var[y_l^{\text{dropout}}]= n_l \beta^2 ( \var[\delta_l]+\ex[\delta_l]^2) \var[w_l] \var[y_{l-1}] \end{equation}

For cascaded $l$ layers and with the fact that $\var[y_1]$ represents the variance of the input layer, we arrive at: 
\begin{equation}
\var[y_L^{\text{dropout}}] = \var[y_1] (\prod_{l=2}^L \beta^2 n_l  (\var[\delta_l]+\ex[\delta_l]^2)  \var[w_l]) 
\end{equation}

For a Bernoulli random variable the maximum $\ex[\delta_l]=\frac{1}{2}$ with $\var[\delta_l]=\frac{1}{4}$, then we can upper bound the $\var[y_L^{\text{dropout}}]$ to be: 
\begin{equation}
\var[y_L^{\text{dropout}}] \leq \var[y_1] (\prod_{l=2}^L \beta^2 n_l  (\frac{1}{4}+(\frac{1}{2})^2)  \var[w_l]) 
\end{equation}

\begin{equation}
\var[y_L^{\text{dropout}}] \leq \var[y_1] (\prod_{l=2}^L \beta^2 n_l  \frac{1}{2}  \var[w_l]) 
\end{equation}

\section{Maxout variance upper bound}
\label{app:maxupper}
$$\textbf{y}_l^{\text{maxout}}= \max\{\textbf{w}_l^i \textbf{x}_l\}^m_i $$
Assumptions: 
\begin{itemize}
    \item $\textbf{w}_l^i  =\{w^i_{kj}: k\in n, j \in d\} \in \mathbb{R}^{n\times d}$ elements are initialized to be mutually independent with zero mean and a symmetric distribution
    \item $\textbf{x}_l =\{x_i: i \in d\} \in \mathbb{R}^{d\times1}$ elements are initialized to be mutually independent sharing the same distribution with zero mean
    \item $\textbf{w}_l$ and $\textbf{x}_l$ are independent from each other

    \item $\textbf{y}_l^{\text{maxout}} =\{y_i: i \in n\} \in \mathbb{R}^{n\times1}$ 
\end{itemize}

The variance of each $y_l^{\text{maxout}} \in \textbf{y}_l^{\text{maxout}} $, where $l$ is the layer number, is: 
\begin{equation}\var[y_l^{\text{maxout}}]= n_l \var [\max\{w_l^i x_l\}^m_i] \end{equation}
This can be upper bounded using the fact the $\var[\max(z_i)_i^m] \leq \var[\sum_{i\in m} z_i]$, applying this bound: 
\begin{equation}\var[y_l^{\text{maxout}}]\leq n_l \var [\sum_{m \in i}\{w_l^i x_l\}]  =   n_l \sum_{m \in i}\var[w_l^i x_l] \end{equation}
Because $w_l^i$ and $x_l$ are both independent and by using lemma~\ref{app:poirv}:

\begin{equation}\var[y_l^{\text{maxout}}]\leq n_l \sum_{m \in i} [ (\var[w^i_l]+\ex[w^i_l]^2)(\var[x_l]+\ex[x_l]^2)  - (\ex[w^i_l]^2) (\ex[x_l]^2)] \end{equation}
\begin{equation}\quad\quad\quad\quad\quad=  n_l \sum_{m \in i} [ (\var[w^i_l]+ \cancelto{0}{\ex[w^i_l]^2})(\var[x_l]+ \cancelto{0}{\ex[x_l]^2})  -  (\cancelto{0}{\ex[w^i_l]^2}) (\cancelto{0}{\ex[x_l]^2})] \end{equation}
\begin{equation}\quad\quad\quad\quad\quad=  n_l \sum_{m \in i} \var[w_l^i] \var[x_l] \end{equation}

By using the the results and assumption made at ~\ref{app:vofnt} given that $x_l=f(y_{l-1})$, we have: \begin{equation}\var[y_l^{\text{maxout}}] \leq  n_l \beta^2 \var[y_{l-1}] \sum_{m \in i}  \var[w_l^i]  \end{equation}

Because all elements of $w^i_l$ share the same distribution, their variance is equal,
we can donate $\var[w^i_l] = \var[w_l]$ and by their independence:
 \begin{equation}\var[y_l^{\text{maxout}}] \leq  n_l m \beta^2 \var[y_{l-1}]  \var[w_l]  \end{equation}

For cascaded $l$ layers and with the fact that $\var[y_1]$ represents the variance of the input layer, we arrive at: 

\begin{equation}
\var[y_L^{\text{maxout}}] \leq \var[y_1] (\prod_{l=2}^L n_l m \beta^2   \var[w_l]) 
\end{equation}

\section{Maxout variance lower bound}
\label{app:maxlower}
$$\textbf{y}_l^{\text{maxout}}= \max\{\textbf{w}_l^i \textbf{x}_l\}^m_i $$
Assumptions: 
\begin{itemize}
    \item $\textbf{w}_l^i  =\{w^i_{kj}: k\in n, j \in d\} \in \mathbb{R}^{n\times d}$ elements are initialized to be mutually independent with zero mean and a symmetric distribution
    \item $\textbf{x}_l =\{x_i: i \in d\} \in \mathbb{R}^{d\times1}$ elements are initialized to be mutually independent sharing the same distribution with zero mean
    \item $\textbf{w}_l$ and $\textbf{x}_l$ are independent from each other
    \item $(w_l^i x_l) \sim \text{iid }\, \mathcal{N}(0,1)$ 

    \item $\textbf{y}_l^{\text{maxout}} =\{y_i: i \in n\} \in \mathbb{R}^{n\times1}$ 
\end{itemize}

From~\cite{ding2015multiple} for $i \in m$ the $\var[\max_i z_i] \geq \frac{c}{\log{m}}$, where  $z_i\sim \mathcal{N}(0,1)$ and $c>0$. Then: 
\begin{equation}
\var[y_l^{\text{maxout}}] \geq n_l \frac{c}{\log{m}}
\end{equation}

\section{Maxout variance asymptotic bound }
\label{app:maxasymp}
$$\textbf{y}_l^{\text{maxout}}= \max\{\textbf{w}_l^i \textbf{x}_l\}^m_i $$
Assumptions: 
\begin{itemize}
    \item $\textbf{w}_l^i  =\{w^i_{kj}: k\in n, j \in d\} \in \mathbb{R}^{n\times d}$ elements are initialized to be mutually independent with zero mean and a symmetric distribution
    \item $\textbf{x}_l =\{x_i: i \in d\} \in \mathbb{R}^{d\times1}$ elements are initialized to be mutually independent sharing the same distribution with zero mean
    \item $\textbf{w}_l$ and $\textbf{x}_l$ are independent from each other
    \item $(w_l^i x_l) \sim \text{iid }\, \mathcal{N}(0,\sigma^2)$ 
    

    \item $\textbf{y}_l^{\text{maxout}} =\{y_i: i \in n\} \in \mathbb{R}^{n\times1}$ 
\end{itemize}

The variance of each $y_l^{\text{maxout}} \in \textbf{y}_l^{\text{maxout}} $, where $l$ is the layer number, is: 
\begin{equation}\var[y_l^{\text{maxout}}]= n_l \var [\max\{w_l^i x_l\}^m_i] \end{equation}

For more details about this proof please check~\cite{2035079,105745,leadbetter1988extremal}.\\

Recall that the standard normal CDF $\Phi$ is such that:
\begin{equation}
1-\Phi(y)\sim\frac{e^{-y^2/2}}{y\sqrt{2\pi}} 
\end{equation}
when $y \to \infty  $ and that, for every $y$:
\begin{equation}
P(Y_n\leq \mu+y\sigma)=\Phi(y)^m
\end{equation}
Hence, if $y_m$ is chosen such that:
\begin{equation}
m\frac{e^{-y_n^2/2}}{y_n\sqrt{2\pi}}=t
\end{equation}
for some given $t$, then: 
\begin{equation}
P(Y_m\leq \mu+y_m\sigma) \to e^{-t}
\end{equation}
Solving this for $y_m$ yields: 
\begin{equation}
y_m=\sqrt{2\ln m}-\frac{\ln\ln m+2\ln t+\ln(4\pi)}{2\sqrt{2\ln m}}
\end{equation}
Then, we can define:
\begin{equation}
\label{eq:varthingy}
Z_m=\sqrt{2\ln m}\frac{Y_m-\mu}\sigma-2\ln m+\frac12\ln\ln m+\frac12\ln(4\pi)
\end{equation}
where, for every real $z$:
\begin{equation}
P(Z\leq z)=\exp(-e^{-z}) = G(\mu=0,\beta=1) \text{ a standard Gumbel distribution.}
\end{equation}
Which aligns with the convergence from extreme value theorem.
From Equation~\ref{eq:varthingy}, for $\mu=0$, we have and for sufficiently large $m$: 
\begin{equation}
   Y_m \approx \frac{\sigma}{\sqrt{2\ln m}}[Z_m+ 2\ln m-\frac12\ln\ln m-\frac12\ln(4\pi)]
\end{equation}
Hence:
\begin{equation}
   \var[Y_m] \approx \var[\frac{\sigma}{\sqrt{2\ln m}}[Z_m+ 2\ln m-\frac12\ln\ln m-\frac12\ln(4\pi)]]
\end{equation}
\begin{equation}
   \var[Y_m] \approx \frac{\sigma^2}{2\ln m}\var[Z_m]
\end{equation}
A standard Gumbel distribution have a variance of $\frac{\pi^2}{6}$, then: 
\begin{equation}
   \var[Y_m] \approx  \frac{\pi^2}{6} \frac{\sigma^2}{2\ln m}
\end{equation}
In our case $Y_m = y_l^{\text{maxout}}$ and $\var[w_l x_l] = \sigma^2$, yielding:
\begin{equation}
\var[y_l^{\text{maxout}}] \approx n_l \frac{\pi^2}{6} \frac{\var[w_l x_l]}{2 \ln m} = n_l \frac{\pi^2}{6} \frac{\sigma^2}{2 \ln m}
\end{equation}

\section{Models training time and parameters count}
\label{app:traintimemodelsize}

\begin{table}[h]
\centering
\scriptsize
\caption{Model parameters count / Average training time per epoch in seconds. M stands for Million, K stands for Thousand. All models were trained on Nvidia V100 GPU. Models in \textcolor{blue}{blue} have the same size at testing time. \textit{+FC} stands for IEN applied on both FC and CNN layers. Maxout results are only on CNN layers. All models have $m=4$}.
\label{tb:parameterssizetime}
\begin{tabular}{llllllll}
Dataset &  & ResNet56 & ResNet110 & DenseNet40-12 & DenseNet100-12 & VGG16 & VGG19\\
\hline
\multirow{4}{*}{CIFAR-10} & \textcolor{blue}{IEN} & \begin{tabular}[c]{@{}l@{}}4.25M / 55.16\\\end{tabular} & 8.60M / 97.17 & 854.94K / 102.79 & 3.74M / 239.49 & 74.11M / 26.62 & 100.66M / 32.34\\
 & \textcolor{blue}{IEN+FC} & 4.25M / 55.9 & 8.61M / 105.38 & 860.26K / 101.01 & 3.75M / 244.74 & 76.23M / 27.46 & 102.79M / 32.93\\
\cline{2-2}
 & Maxout & 4.25M / 57.45 & 8.60M / 110.53 & 854.94K / 106.33& 3.74M / 244.74 & 74.11M / 28.20 & 100.66M / 33.98\\
\cline{2-2}
 & \textcolor{blue}{Base} & 853.02K / 17.16 & 1.73M / 42.01 & 176.12K / 51.96 & 769.16K / 115.16 & 15.25M / 7.61 & 20.57M / 9.21\\
\hline
\multirow{4}{*}{CIFAR-100} & \textcolor{blue}{IEN} & 4.25M / 58.56 &8.61M / 94.75 & 866.91K / 96.13 & 3.77M / 239.63 & 74.16M / 26.68 & 100.71M / 32.29\\
 & \begin{tabular}[c]{@{}l@{}}\textcolor{blue}{IEN+FC}\\\end{tabular} & 4.28M / 58.30 &8.64M / 96.38 & 920.11K / 98.71 & 3.75M / 244.74 & 76.46M / 27.32 & 103.02M / 33.10\\
\cline{2-2}
 & Maxout & 4.25M / 62.64 & 8.61M / 111.1 &866.91K / 104.53 & 3.77M / 253.85 & 74.16M / 28.18 & 100.71M / 34.02\\
\cline{2-2}
 & \textcolor{blue}{Base} & 858.87k / 20.01 & 1.73M / 36.28 & 188.09K / 47.73 & 800.3K / 114.13 & 15.30M / 7.62 & 20.61M / 9.20\\
\bottomrule
\end{tabular}
\end{table}

From Table~\ref{tb:parameterssizetime} we notice that Maxout training time is always more than IEN. There is also an increase in both training time and parameters size of IEN+FC against IEN only. Training time of IEN with $m=4$ is much cheaper than training $4$ ordinary models, but it comes at the cost of the memory only during the training time.

\section{Empirical analysis of dropout}
\label{app:emprdropout}
\begin{table}[h]
\centering
\caption{Results of applying dropout everywhere except the input and output layer. Mean error rate and std for multiple runs are reported. The lower the better. IEN uses $m=4$.}
\label{tb:dropout}
\begin{tabular}{llllll}
Dataset &  & IEN & Base & Dropout & Dropout+IEN\\
\hline
\multirow{2}{*}{CIFAR-10} & \begin{tabular}[c]{@{}l@{}}\textcolor{black}{ResNet56}\\\end{tabular} & \begin{tabular}[c]{@{}l@{}}6.93$\pm$0.26\\\end{tabular} &8.38$\pm$1.2 & 90.48$\pm$0.04 & 90.12$\pm$0.01\\
 & \textcolor{black}{ResNet110} & 6.27$\pm$0.32 & 6.38$\pm$0.48 & 91.29$\pm$0.07 & 90.13$\pm$0.01\\
\hline
\multirow{2}{*}{CIFAR-100} & \begin{tabular}[c]{@{}l@{}}\textcolor{black}{ResNet56}\\\end{tabular} & 29.34$\pm$0.49 & 29.97$\pm$0.71 & 99.09$\pm$0.015 & 98.98$\pm$0.01\\
 & \begin{tabular}[c]{@{}l@{}}\textcolor{black}{ResNet110}\\\end{tabular} & 28.17$\pm$0.11 & 27.83$\pm$0.64 & 98.91$\pm$0.01 & 98.98$\pm$0.01\\
\hline
\end{tabular}
\end{table}

We wanted to analyze the effect of using dropout everywhere except the input and output layers. Dropout performance is the best when its probability is set to 0.5. We tested two combinations, dropout applied as before and IEN followed by dropout. Table~\ref{tb:dropout} summarizes these experiments. We notice that dropout only leads to very poor performance in comparison with IEN and base models. This shows that even though theoretically dropout can lower the variance, it is not applicable everywhere unlike IEN. Also, from the same Table~\ref{tb:dropout} we notice that when IEN is combined with dropout the results get slightly better. This suggests the power of IEN in reducing the variance even when the model is being driven by another component.
\section{Guide to the code}
\label{app:code}
Most of the models needed 1x Nvidia V100 with 32 GB of GPU memory for training. Some DenseNet models needed 2x Nvidia V100. Our code is attached for results reproduction with a quick README.MD.

\subsection{ResNet Details}
We used an open source repository~\cite{Idelbayev18a} that produces the same results like ResNet original paper using Pytorch~\cite{paszke2019pytorch}. The models were trained for 200 epochs using a learning rate of 0.1 and batch size of 128. 

\subsection{DenseNet Details}
We used an open source repository~\cite{pleiss2017memory} that produces the same results like original DenseNet paper but in more memory efficient way using Pytorch. The models were trained for 300 epochs using a learning rate of 0.1 and batch size of 64. The is code located at:\\

\subsection{VGG Details}
We used an open source repository~\cite{ChengYangFu}. The models were trained for 300 epochs using a learning rate of 0.9 and batch size of 128. 

\subsection{NAS Details}
We used the default settings in NAS-Bench-201~\cite{dong2020nasbench201}. We only added our IEN and Maxout operators and searched, trained the models. 

\end{document}